# Advanced Machine Learning Framework for Efficient Plant Disease Prediction

Aswath M
Dept. of Electronics and
Communication Engineering,
VIT,
Chennai, India,
m.aswath08@gmail.com

Sowdeshwar S
Dept. of Electronics and
Communication Engineering,
VIT,
Chennai, India,
sowdeshstudies@gmail.com

Saravanan M
Ericsson India Global Services
Pvt. Ltd., Ericsson Research,
Chennai, India,
m.saravanan@ericsson.com

Satheesh K Perepu
Ericsson India Global Services
Pvt. Ltd., Ericsson Research,
Chennai, India,
perepu.satheesh.kumar@ericsson.com

*Abstract*—Recently, Machine Learning (ML) methods are built-in as an important component in many smart agriculture platforms. In this paper, we explore the new combination of advanced ML methods for creating a smart agriculture platform where farmers could reach out for assistance from the public, or a closed circle of experts. Specifically, we focus on an easy way to assist the farmers in understanding plant diseases where the farmers can get help to solve the issues from the members of the community. The proposed system utilizes deep learning techniques for identifying the disease of the plant from the affected image, which acts as an initial identifier. Further, Natural Language Processing techniques are employed for ranking the solutions posted by the user community. In this paper, a message channel is built on top of Twitter, a popular social media platform to establish proper communication among farmers. Since the effect of the solutions can differ based on various other parameters, we extend the use of the concept drift approach and come up with a good solution and propose it to the farmer. We tested the proposed framework on the benchmark dataset, and it produces accurate and reliable results.

## I. INTRODUCTION

Agriculture is the fundamental building block of a society. It is important to address the problems faced by farmers and provide them easy ways to follow necessary steps with an advanced technical application [1]. When there is not much technical assistance available in the local area, farmers can leverage the online platform to post their concerns and receive feedback. However, there is an issue of filtering out trusted individuals who give possible responses from the wrong ones. To manage this verification of users, we make use of a virtual bot account. This bot establishes a platform for the farmer's tweets. The users' solutions to the farmer's queries are verified by the bot before it is considered for evaluation. After that, the bot replies to the responses to the farmer's original tweet.

Lack of advanced equipment and delay in the identification of plant diseases affect the quantity and quality of the crop yield to a great extent. According to studies, the losses due to plant diseases account for 20 to 40 percent of global annual productivity [2]. Extensive yield loss also contributes to various other factors like increased consumer prices which causes a bump in the earnings for the consumers. Hence, the need for timely identification of plant diseases will play a vital role in ensuring good yield which is highly important for farmers and crop producers in remote areas as they cannot afford a loss in the present competitive scenario. The advancements in machine learning algorithms especially in image processing techniques made it possible to perform disease identification and these can be used in real-time to perform plant disease detection, which can save both time and cost [3]. As technology is developing every day, the devices have become cheaper with better resolution, good picture quality, and so on. Here, image processing has become very precise and effective as well. In this paper, we mainly aim at detecting various plant diseases by applying suitable image processing techniques and providing solutions based on the users' inputs in social media. We found that these solutions are differ based on the seasonal variations.

The proposed work provides an approach to design a system that not only helps the user to detect various diseases, but it also provides a real-time solution for that disease based on the suggestions from the people on social media. We also aim at providing the best solution to the user by considering the season and various other factors. This work can be transformed into a mobile application so that the task of disease identification through visual classification can be done in a single approach. The main objectives of the work are to identify whether a leaf is diseased or healthy and classify the type of the disease and it can be accomplished in scenarios where the plant disease is visually distinguishable. The previous research in this domain is intended to identify the diseases from a few crops or diseases with small sample size and limited set of rules [4]. Therefore, they provide a solution of restricted utility, and the solutions for the diseases were not fully ascertained. In this paper, a social media-based approach is proposed which tries to overcome the limitations of previous work in terms of scalability and portability. Hence the essence of the proposed research lies in the provision of high accuracy in performance and portability to mobile devices. Here, the image of the disease is posted on social media, and various solutions and allied images provided by the people are collected in this process. The images provided by the people are then given as input to the model to verify whether the name of the disease and the image provided were match with each other and it includes the consideration of the seasonal impacts. The solutions given by the people are prioritized and finally, output is provided to the user. Now we will discuss the use of advanced machine learning models to address the specific challenges in the paper.

## A. Neural Network Based Image Analysis

Neural networks possess the behavior of a human brain, allowing computers to recognize and solve the problems in the field of artificial intelligence, machine learning, and deep learning [5]. Artificial Neural Networks consist of various layers of interconnected artificial neurons powered by activation functions that help in firing them. But unlike the human brain, which is associated with various activities simultaneously, the efficiency of the neural networks depends on the algorithm used and the number of tasks assigned to them. In this paper, we have employed the Convolutional Neural Networks (CNN's) [6] for image processing tasks to understand and classify the plant images by considering the disease as an indicator. In addition to the image processing, one may investigate the changes happening in the data due to various reasons.

## B. Concept Drift – Determine the Changing Scenario

In the real world, concepts are often not stable but change with time. Typical samples of this are weather prediction rules and customers' preferences. The underlying data distribution may change as well. Often these changes make the model built on old data inconsistent with the new data, and hence it is important to update the model regularly. This problem, referred to as concept drift, complicates the task of learning a model from data and requires special approaches, different from commonly used techniques. It treats arriving instances as equally important contributors to the final concept. This paper considers different types of concept drift, peculiarities of the matter, and provides a review of existing approaches to the matter. Two kinds of concept drift that will occur in a real-world scenario are generally provided in the literature [7,8,9]: (1) sudden (abrupt, instantaneous), and (2) gradual. For example, someone graduating from college might suddenly have completely different monetary concerns, whereas a slowly wearing piece of factory equipment might cause a gradual change in the quality of output parts. Stanley [7] divides the gradual drift further into moderate and slow drifts, counting on the speed of the changes. Hidden changes in context might not only be an explanation for a change of target concept, but it can also cause a change in the underlying data distribution. Even if the target concept remains an equivalent, and it's only the information distribution that changes, this might often cause the need of revising the present model, as the model's error may not be acceptable with the new data distribution. The necessity within the change of current model recognizes the change of knowledge distribution which is named as virtual concept drift [8]. Virtual concept drift and real concept drift often occur together. Virtual concept drift alone may occur in the case of spam categorization as an example. While our understanding of an unwanted message may remain equivalent over a comparatively long period, the frequency of various sorts of spam may change drastically with time. In [9] virtual concept drift is mentioned as sampling shift, and real concept drift is mentioned as concept shift. In our problem, the type of concept drift is ascertained by considering the seasonal benefits and customer preference in the proposed model. We tried to explore all the changes in a common social media platform where people can communicate regularly.

## C. Social Media and Other Communication with NLP

Nowadays, people are using social media platforms extensively in their day-to-day life. A lot of information is shared in just a fraction of a millisecond. In such a fast-moving and more connected world, opportunities for development related to the society are certainly made possible. We leverage this ideology and build on top of a buzz-filled social media platform to help farmers healthily accelerate their crop production. This also provides society to witness the kinds of problems that farmers face. Textual information is processed with Natural Language Processing (NLP) techniques [10]. The farmer posts a tweet about the problem that he is facing along with an image of the diseased plant. The Twitter community suggests solutions that the farmer could try to solve the problem. These suggestions are extracted, and the text is processed with NLP methods. There are quite a good number of tools and techniques in NLP for processing text data. The fundamental method of parsing textual data is word embeddings. Each word is associated with a vector that contains scalar quantities that denotes its proximity with other words. More advanced methods of parsing text include CNN-based approaches. In this paper, we have explored Word Mover's Distance (WMD) [11] to find the similarity between the user tweets and the trusted database solutions. WMD uses word embeddings as a basis to work on. There are word embedding techniques like Word2Vec [12]and Glove [13] which give the substantial quality of vector representations and this is leveraged in WMD [11] itself. The main aspect of WMD is its ability to deal with a string of words in a sentence as opposed to working with word-by-word comprehension. The distance between two documents is calculated as the sum of minimum distances that a word from document A proceeds to reach document B.

## II. RELATED WORKS

Nowadays smart agriculture gained recent attention and many people have proposed different automation mechanisms in it [1,2,3]. In [1], the authors proposed a platform to detect plant diseases and obtain remedies by taking assistance from the people on the Twitter platform. However, there are a couple of problems with this approach (i) the dataset used in the paper is very small and hence we cannot obtain a good model, and (ii) the approach used in the paper cannot used in all the conditions as remedies can change depending on the season and other factors. In another work, authors proposed a reference architecture for integrated farming systems which provides a comprehensive overview of a high-level integration of various technologies and their interplay is described [3].

The image-based assessment approaches are proven to produce more accurate and reliable results than those obtained by human visual assessments. Stewart and McDonald [14] used an automated image analysis method to inspect disease symptoms of infected wheat leaves caused by Zymoseptoria tritici (a plant disease). This method enabled the enhancement of pycnidia size and density, along with other traits and their correlation, which provided greater accuracy and precision compared with human visual estimates.

In [15] CNN has been used to detect diseases in crops with good accuracy measures. Muhammad Hammad Saleem et. al [16] addressed plant disease detection and classification by developing deep learning architectures along with several techniques to detect and classify the symptoms of plant diseases. In [16], the authors have used a public dataset known as Plant Village which contains about 54,306 images of diseased and healthy plant leaves. We have used this public dataset for our model training. In [17] they have trained a deep neural network model that identifies 14 crop species and 26 diseases with very good training accuracy. The architecture used here is GoogleNet [18] and the transfer learning mechanism is also explored.

All these existing studies deal with detecting plant diseases only. However, in addition, we need to aid the farmers on the remedies to solve these problems as an end-to-end system. Hence, in this paper, we have trained a new model to provide solutions to the diseases by retrieving the replies from the people after two levels of filtering. Moreover, these solutions are dynamic and dependent on other factors. The solution for the same disease in different seasons might be different. So, appropriate solutions are to be given to the user by considering all the factors. In literature, people call it concept drift [19], which deals with change in the data distribution.

As mentioned in the work [20], dissimilarity measure is still an open-ended question to detect the concept drift. In this paper, we propose a new method to measure this dissimilarity between the old data and new data, using Hamming distance. The concept drift and its applications are briefly explained in [8, 9]. Even if the solutions given by the people are correct, it might not be an appropriate solution for that season. So, we have applied concept drift techniques to handle this problem. Moreover, [20, 21] explain the advantages of using multiple models instead of a single complicated model. Using this information and results, we have designed two different models for different seasons. The specific model to use from all these models is selected by using the Hamming distance metric. In this way, even though there are multiple solutions for that disease, an appropriate solution is selected and displayed.

III. PROPOSED APPROACH

In this section, we explain the proposed approach to construct the digital platform along with the different components. The architecture of the proposed approach is shown in Fig. 1. Our novel architecture encapsulates the communication and data flow of the original tweet text, deep learning inference, NLP text processing, and finally replying the solution to the original farmer's tweet. Our model can detect images of diseased leaves under varying lighting conditions, noisy images, and even with stretched aspect ratios. The NLP model is used to perform multiple tasks such as verification of the user's solution before they are taken for final evaluation. Additionally, in periodic moments, the data is checked for detecting concept drift.

The process starts with the user posting the tweet on the help needed along with the image of the plant with a unique hashtag. For this tweet, we ask users to reply with the disease name and remedy details using the same hashtag. Now, we will follow the below steps to identify the plant disease and come with a top remedy for the identified disease. The entire process is described in Fig. 1.

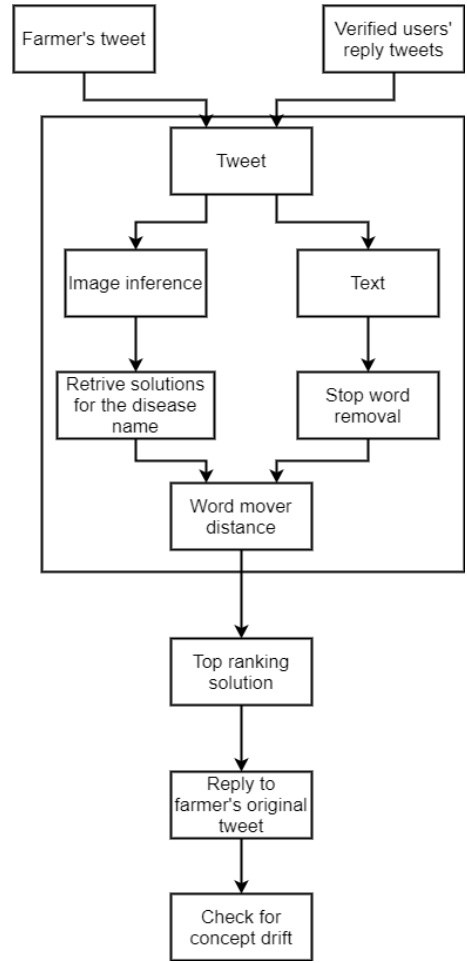

Fig. 1. Flow diagram of proposed plan disease detection system.

First, we filter the tweets based on image classification techniques to match the images with that of a trusted database (Plant Village).

Second, we filter the remaining tweets using NLP techniques and rank these solutions.

Finally, we use the concept drift approach to use the right model to arrive at the best solution.

In Fig. 2, we included all the components in a step-by-step fashion. Following are the steps we perform in the proposed approach to detect the plant disease and the solution identification.

1. The user posts the image of the plant with the suspected disease along with the information on other sources like sensor readings with a particular hashtag.

2. The user who sees the tweet replies with the disease name, matching image, and probable solution with the same hashtag.

3. First, the disease name is filtered based on the number of votes by the people in their replies.

4. Next, we filter out the remaining tweets based on the image posted and database image corresponding to the voted disease.

5. Finally, the solution to the disease is given to the end-user by selecting the best prediction model based on the WMD metric.

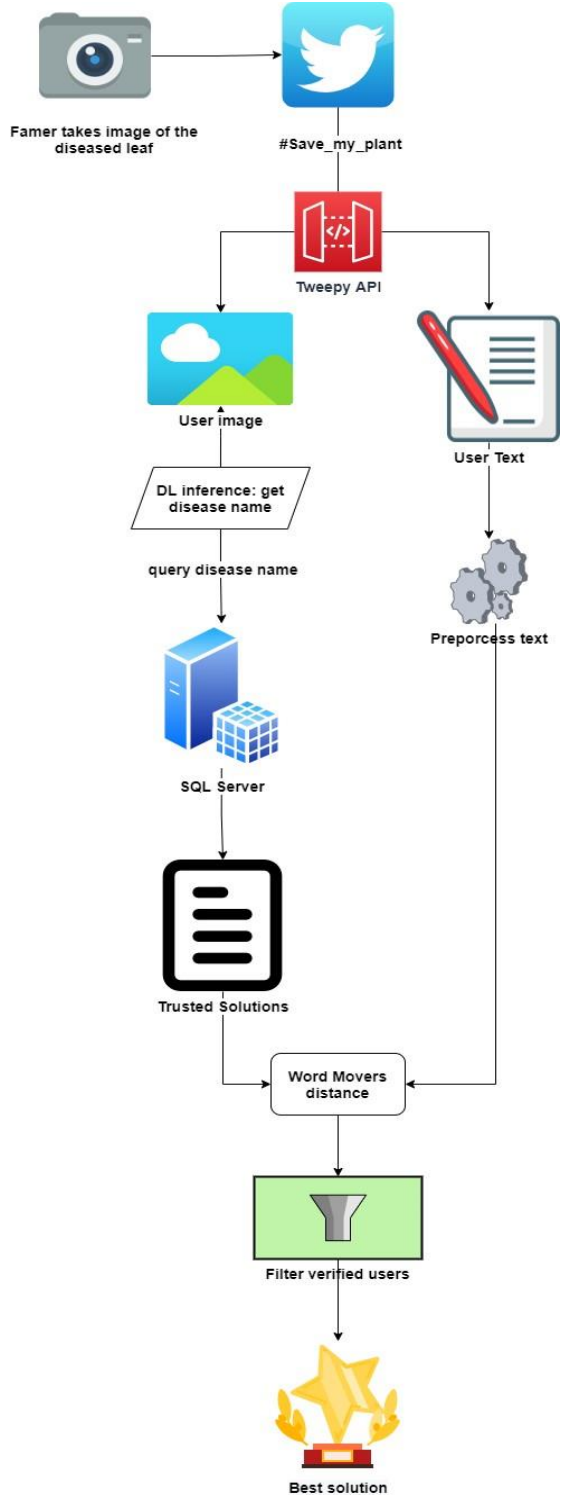

Fig. 2. Architecture of the proposed plant disease detection.

## A. Deep Learning Based Visual Disease Detection – Dataset and Model Building

**Dataset:** The model is trained using the plant village dataset which contains several images of healthy and infected leaves of different classes of plants. This dataset is a crowdsourcing effort to help the farmers or researchers, find the diseases effectively and efficiently. Each image is of size 256 * 256. The different types of plants used was pepper bell, potato, and tomato.

**Model:** We built an Inception-Resnet-v2 model [22] using transfer learning which takes in the images from the plant village dataset and classifies if the plant has any type of disease as well the name of the plant. The model contains CNN layers in the beginning and dense layers at the end. The Softmax function is applied to the output layer to find the appropriate class. Inception-Resnet-v2 is based on the combination of Inception structure and residual connection. In the Inception-Resnet block, multi-size convolution filters are combined by residual connections. The use of residual connection avoids the degradation problem caused by the deep structure. It also reduces the time. This architecture is highly efficient and when trained at 25 epochs we were able to achieve the maximum. Fig. 3 shows the basic architecture of the Inception-Resnet-v2 network.

**Training process:** We trained the model for 30 epochs with a batch size of 32 images. The dataset has 15 classes for classification tasks. We used Adam's optimizer [23] with an initial learning rate of 0.001. Binary cross entropy measure was used as the metric of loss on which the model is trained.

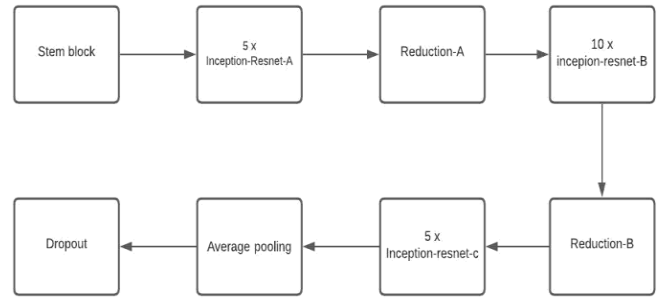

Fig. 3. The basic architecture of Inception-Resnet-v2.

The architecture of the model architecture used in this work is shown in Fig. 4 and 5. The model contains convolutional layers followed by pooling layers and dropout layers. Also, to prevent overfitting we used batch normalization layers in the middle.

## B. Application of NLP in Text Processing

A curated list of the most prevalent diseases is taken and entered in a comma-separated value file. The corresponding solutions to the disease names are recorded in the adjacent column. The CSV file is stored in the SQL database for ease of access. To implement on the social networking system, we extract the tweets using the Tweepy library [24] and various NLP techniques were used to extract useful information from the extracted tweet messages. First, the tweets are posted by users in a predetermined format. The format requires two labels "name" and "solution" followed by a colon. Also, it is required that the tweet is posted with a particular hashtag, in our case it was #savemyplant. Tweepy Python API is used to search and retrieve the tweets containing the used hashtag for further processing. Using regular expressions, the name of the disease and the suggested solution by the users are extracted. The extracted names and solutions are then passed to the pre-processing stage. In the pre- processing stage, we perform various operations for the extracted text. These techniques include stop word removal, punctuation removal, and trimming of white space characters. Stop word includes articles, pronouns, and prepositions. The collection of stop words is provided by the NLTK library. The preprocessed disease name is taken and matched with the closest names registered in the trusted database.

The trusted database is the one that is supplied by the Government/verified Private company to ensure no wrong solutions are given to the Farmer. In this work, we consider the trusted source as solutions available in the plant village dataset. In the future, we change the trusted database to one given by the Government/verified private companies. The trusted solutions corresponding to the disease name are retrieved with a SQL "where" query.

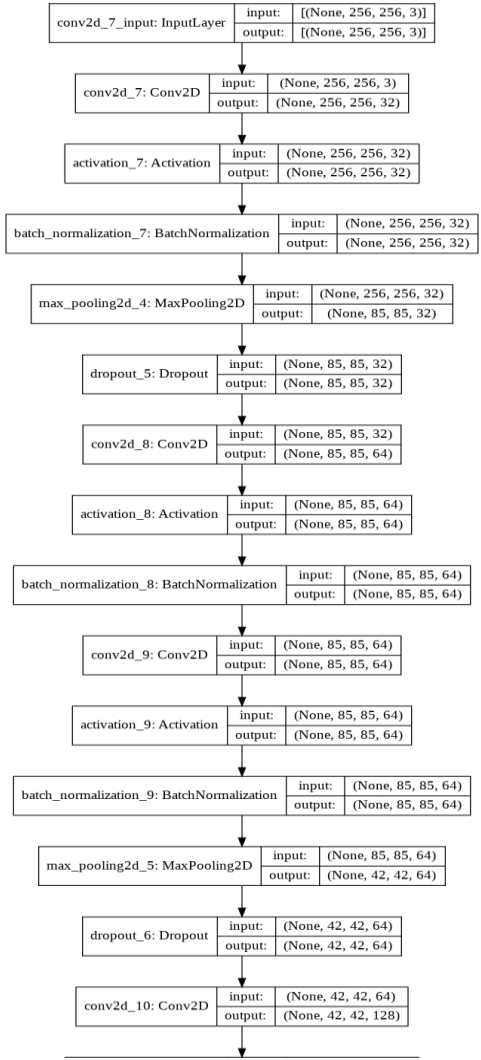

Fig. 4. Model Part 1.

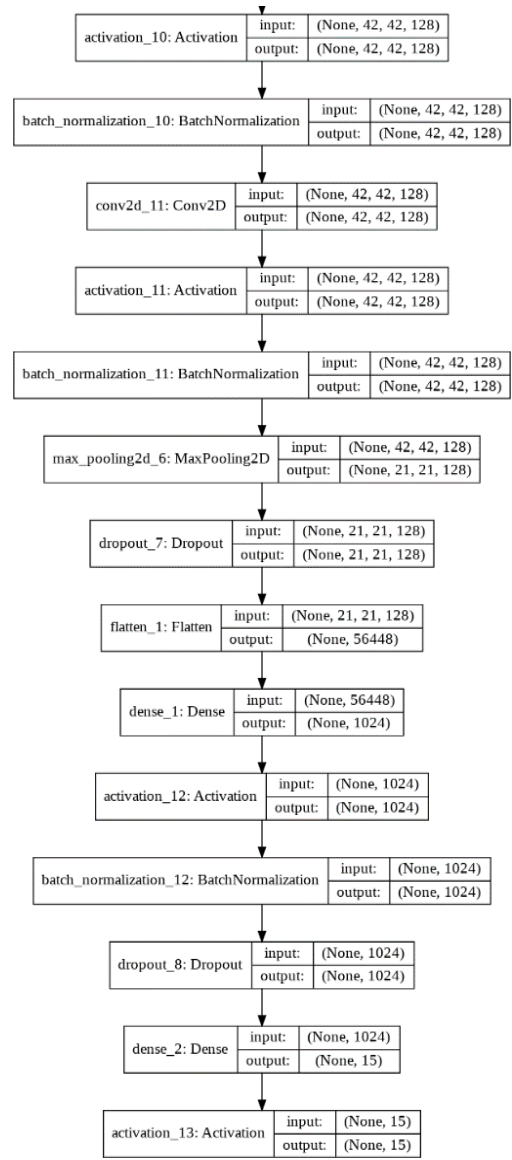

Fig. 5. Model Part 2.

*C. Concept Drift*

Concept drift method is used for switching the prediction model. Each season will have different effects on plants. And plants also respond differently to remedies under different seasons. The exact time when a new season might arrive would be quite unpredictable. Because of this, we could switch to a different model when the next season arrives.

There are two types of drifts: Data drift and Concept drift. In Data drift, the data that is originating from the source is drifting away, from the regular cycle. In concept drift, the suggested solutions by the community are deviating from the trusted database solutions. Hence, about our application, Data drift and Concept drifts can be given the above-discussed definitions.

A resilient disease detection model would adapt to the concept, one of the methods is based on the understanding of past data and trends. The model would evaluate the conditions based on Bayesian beliefs. In the standard Bayesian model, with knowledge of a prior belief and the likelihood of an event to occur, the posterior belief can be determined. If the posterior belief crosses a predetermined threshold, then the model switch could be performed. Cloud servers are used to perform service to millions of people. With some prior knowledge of the situation and a smaller model, the number of computations needed to run inference on the model can be vastly reduced. Thus, making the servers efficient and responsive. The implementation of concept drift is explained as shown in equation (1).

$$P(S|E) = \frac{P(E|S)P(S)}{P(E|S)P(S)+\sum_{i=1}^{N} P(E|H_i)P(H_i)} \quad (1)$$

The terms in equation (1) are explained as follows:

- $N$ is the total number of seasons on which models are trained. If more seasons are identified, then N increases. For each season $N$, we have a CSV (Comma Separated Value) file containing details of the diseases such as the disease name and the solutions, water availability, number of daylight hours in a given region, dangerous pests, and the time of the year this event happens.

- $E$ is an event that happens at any given time in a year. An event contains information in text form. The text contains details regarding the current season, identified disease names, and solutions.

- $P(S|E)$ is the probability of season S to occur given that event E has occurred. A Drift is detected as the probability of this term changes. The lower this probability gets, the more probable of switch to a new model.

- $P(S)$ is the overall probability of season S to occur. This can be defined as a fraction of the period in months that season S is likely to occur in a year based on known historical data. Season S is one among the N seasons.

- $P(H_i)$ is the probability of season $H_i$ to occur based on historical data. Seasons $H_i$ is all the N seasons excluding season S. These seasons are expected to occur once in a year, however, the exact time of the year at which it occurs cannot be precisely predicted. Only the approximate duration of how long the season $H_i$ lasts is known. The probability $P(H_i)$ is the ratio of this duration to the entire year.

- $P(E/S)$ is the probability of event E occurring in season S. This probability is evaluated by running inference on the tweet information with season model S.

- $P(E/H_i)$ is the probability of event E occurring in season $H_i$. This probability is evaluated by running inference on the tweet information with season model $H_i$.

The advantage of using concept drift is that it is easy to train a specific model for each season. Instead of training a large, all-encompassing model, smaller yet, multiple models can be used. Having many different models rather than a single large one means that we can have more low-level control of the whole operation. If we were using a single model, then a lot of computations time and energy would be wasted. In the case of a larger model, the training time may be significantly large as more parameters may be modified. However, If one of the models is not performing well on a particular season, then that particular model can be taken and retrained to increase the accuracy. This reduces the model training time. Moreover, in the real-time deployment of these models, by using concept drift method, the number of computations needed for running inference can be reduced greatly. This can be a benefit as the cloud servers can serve faster and with less latency. Fig 6 shows the overview of implemented approach of concept drift in this paper.

The next step is to perform the dissimilarity measure of the drift data that is accumulated over time. A sliding window of last $R$ results of $P(S|E)$ are stored in a 1D array of length $R$. Ideally, Linear least squares can be used to fit a line through set of data points. However, the Linear least squares method does not perform well if there exists even a single large outlier. Hence, we try to use another alternative for approximating statistical difference known as RANSAC (Random Sample Consensus) [24] which can be used to find the best-fit slope of the drift. If the slope over the entire range of $R$ is significantly large, then the model can be switched. The problem in RANSAC lies in its random sampling and trial-based approach to converge on the solution.

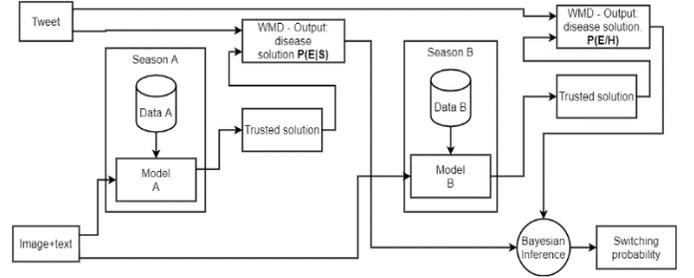

Fig. 6. Concept drift.

In RANSAC, the number of trails need to perform to converge to the best solution can be computed statistically as shown in equation (2). We start by defining $N$, which is the maximum trails that are required to run to obtain the best fit of a given confidence level. Here $k$ is the number of samples from the total number of data points chosen to build a model, where $k$ ranges from 0 to $N$. The confidence level $p$ of the solution whose value ranges from 0 to 1. The term $e$ is the ratio of the number of inliers to the number of samples at a given trial is given. If the slope obtained from the drift data-point is large, then the current model can be replaced by a better matching model.

$$N = log(1 - p)/log(1 - e^k) \quad (2)$$

IV. EVALUATION AND DISCUSSION

The techniques discussed in section 3 are evaluated and the metrics are presented in the following subsections. Starting from the accuracy of leaf image-based disease detection, the performance of NLP techniques, Concept drift, and Social media.

## A. Disease Detection from Leaf Images

The Deep Learning model used to detect disease names with the images of leaves gives the accuracy of 99.30% percent training accuracy and 98.77% percent validation accuracy, the metrics of the training are shown in Fig 7, the model was trained with a dataset containing 10 different classes of diseases.

The prediction step was running on an 8-core AMD Ryzen processor equipped Nvidia RTX 2060 GPU takes around 0.8 seconds. This result compares better with our previous implementations [15]. In our previous implementation, the custom-built CNN model gave an accuracy of 93.1% for correctly classifying a label. The SVM algorithm as discussed in [1] gave an accuracy of 79.97%. With every epoch, the training and validation accuracy increases, particularly, at epoch 8, there is a sharp jump when only the last 70 layers are made trainable, and the rest of the model's head is frozen. Moreover, the second half of Figure 7 shows the graph of decreasing of cross-entropy loss with logits on both training and validation data.

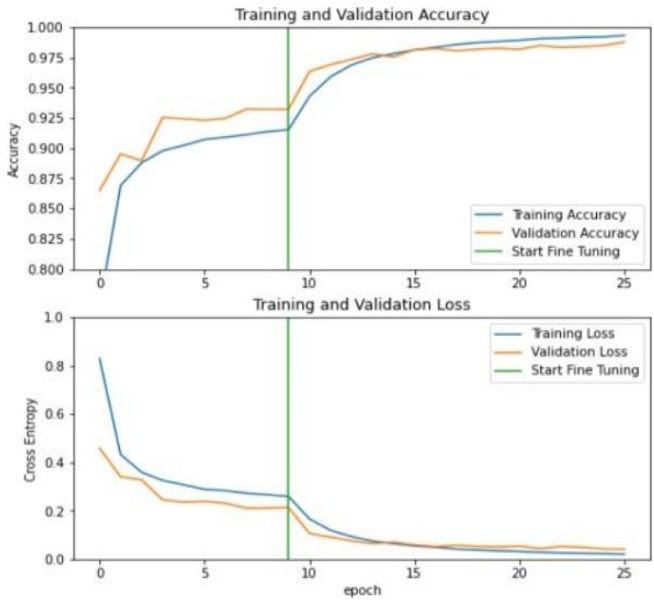

Fig. 7. InceptionResnetV2 metrics.

## B. WMD Implementation

In this framework, we used the WMD metric to compare two texts coming from two different models. The WMD metric can help us to calculate similarity among texts even when the words are not exact match i.e., it can help us to find the similarity among synonyms. The idea is to compute the word embeddings and then compute the distance between them. For example, let us take two sentences 'Moisture the soil' and 'Add water to the plant'. Now, although the two sentences are different the underlying meaning is the same. The WMD metric can help us in achieving the similarity between these types of sentences as users can answer with different words although the underlying meaning can be the same.

To do this we first pre-process the data by removing the stop words, lemmatization, etc. Further, on the remaining words, we use sentence embeddings to convert the sentence to a numerical vector. Finally, the obtained word embeddings are normalized, and distance is calculated. Essentially the WMD will choose the minimum transportation cost to transport every word from two predictions obtained from models and benchmark text. Depending on the WMD metric obtained, we flag the output as concept drift and decide which model to use. More details of its usage are discussed in the next sub-section.

## C. Understanding Seasonal Drift

Seasonal drift is an important thing to arrive at a good solution. For example, to handle a disease 'X' we can have two solutions depending on the season. In the summer season, we need to use a solution as 'Water the plant'. The same solution may not be valid in the rainy season since watering may not be essential during the season.

As per eqn. (1), the event is the solution name 'Water the plant' and season in summer. Now, we will calculate the $P(S|E)$ for the rainy season and if the value is low, then we can say with high confidence we cannot use the same solution and we need to use another model to predict the solution. In this way we can detect the concept drift in the solution and able to select the best model for the given solution.

Model switching for different seasons also gives desirable results. The probability of switching to a new model is identified correctly by the concept drift algorithm. The gradual change in the deviation as the days goes by is shown in the graph in Fig 8. At every instance of time in a month, as the solutions deviate from the current model's prediction, the WMD value gets larger. Thus, the probability of switching to a new model will increase.

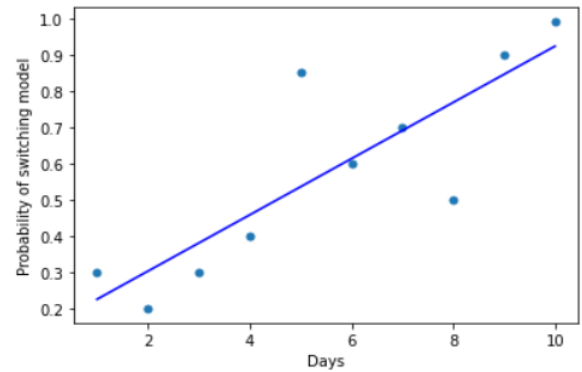

Fig. 8. Concept drift detection with RANSAC Line fitting.

The WMD gives a value in the range of 0 to slightly more than 1. In cases where the WMD is greater than 1, the value is clipped and finally taken as 1 for performing the Bayesian inference [25]. When the WMD of the current model is divided by the value of WMD of a more matching model, the resultant value is much higher, this large value indicates that the model needs to be switched to the more matching one. If the model gives a WMD value of 0.38 on the current model, and 1.2 and 0.1 on two other models, then the effective probability is 0.25. If that same probability is evaluated for the most probable model, the model which gave 0.1, the effective Bayesian inference of switching to another model is 0.067. It means that the model which gives the least probability of switching to another model is chosen as the best model for the next season.

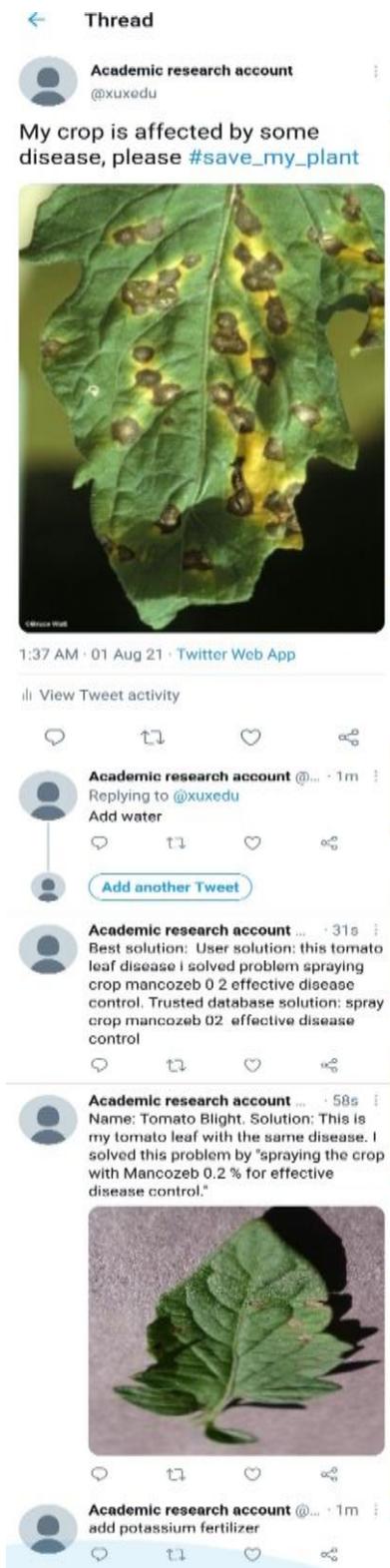

Fig. 9. Interaction between farmers, users, and a Bot in Twitter.

*D. Social Media*

All the above-discussed ideas are finally integrated and made accessible for people with a Twitter social media platform. Fig 9 shows this interaction between the integrated together farmer, users, and the bot. The top of Fig 9 shows that a farmer had posted a tweet with the hashtag #save my plant and attached the image of a diseased leaf from his agriculture field. As we move down in the figure from the top, the public or a trusted circle of people starts providing suggestions in the reply section of the farmer's tweet. In the following replies at 1 minute, two users had replied "Add water" and "add potassium fertilizer". A third user had also tweeted a reply at 58 seconds, containing the name of the disease and a descriptive solution of the same problem that he had faced and the image of his diseased plant leaf. The reply is posted in a predefined format with "Name:" and "Solution:" keywords visible so that our algorithm extracts the information with regular expressions in Python language. This user's reply is given a higher weightage while evaluating the solution because of his descriptive reply. After a while, typically within a day, our algorithm scrapes all the tweets that contain the hashtag #save my plant. The user's reply tweet to the main farmer's tweet is also scraped. First, the image in the farmer's tweet is analyzed by our algorithm with the CNN method, followed by the analysis of the replies. The user replies are compared with a trusted database to rank the best solution. Finally, the "Best solution" containing the "User solution" and the "Trusted database solution" is posted by our bot, as posted at 31 seconds.

V. CONCLUSION

A framework was developed to aid farmers in identifying plant diseases in this paper. The proposed framework is built on top of the Twitter platform and participants from a user community can reply to the initial message posted by the farmer. The framework uses deep learning techniques to classify the images posted by the user community and natural language processing techniques to rank the solutions posted by the users. Since the seasonal and several other parameters can affect the solution, we came up with a method to detect the right solution using the mechanism of concept drift. The proposed framework is tested on the plant village dataset and the solutions are verified with the end-users who are more satisfied with the solution. Future directions will include the testing of the framework on different datasets. Moreover, the solution can be extended to different closed community platforms such as Facebook, WhatsApp, etc.